\documentclass{article}

     \usepackage[preprint]{neurips_2020}

\usepackage[utf8]{inputenc} 
\usepackage[T1]{fontenc}    
\usepackage{hyperref}       
\usepackage{url}            
\usepackage{booktabs}       
\usepackage{amsfonts}       
\usepackage{nicefrac}       
\usepackage{microtype}      
\usepackage[pdftex]{graphicx}
\usepackage{multirow}
\usepackage{amsmath}

\usepackage{algorithm}
\usepackage{algpseudocode}
\title{MultiResolution Attention Extractor for Small Object Detection}

\author{%
 Fan Zhang\thanks{Key Laboratory of Intelligent Perception and Image Understanding of Ministry of Education, International Research Center for Intelligent Perception and Computation, Joint International Research Laboratory of Intelligent Perception and Computation.} \\
  School of Artificial Intelligence\\
  Xidian University\\
  Xian, Shaanxi Province 710071, China \\
  \texttt{zhangfan\_1@stu.xidian.edu.cn} \\
   \And
   Licheng Jiao \\
  School of Artificial Intelligence \\
  Xidian University \\
  Xian, Shaanxi Province 710071, China \\
   \texttt{lchjiao@mail.xidian.edu.cn} \\
  \AND
   Lingling Li \\
  School of Artificial Intelligence \\
  Xidian University \\
  Xian, Shaanxi Province 710071, China \\
  \texttt{linglingxidian@gmail.com} \\
  \And
  Fang Liu \\
  School of Artificial Intelligence \\
  Xidian University \\
  Xian, Shaanxi Province 710071, China \\
  \texttt{f63liu@163.com} \\
  \And
  Xu Liu \\
  School of Artificial Intelligence \\
  Xidian University \\
  Xian, Shaanxi Province 710071, China \\
  \texttt{xuliu361@163.com} \\
}

\begin{document}

\maketitle

\begin{abstract}

Small objects are difficult to detect because of their low resolution and small size. The existing small object detection methods mainly focus on data preprocessing or narrowing the differences between large and small objects. Inspired by human vision “attention” mechanism, we exploit two feature extraction methods to mine the most useful information of small objects. Both methods are based on multiresolution feature extraction. We initially design and explore the soft attention method, but we find that its convergence speed is slow. Then we present the second method, an attention-based feature interaction method, called a MultiResolution Attention Extractor (MRAE), showing significant improvement as a generic feature extractor in small object detection. After each building block in the vanilla feature extractor, we append a small network to generate attention weights followed by a weighted-sum operation to get the final attention maps. Our attention-based feature extractor is ${2.0}\times$ the AP of the “hard” attention counterpart (plain architecture) on the COCO small object detection benchmark, proving that MRAE can capture useful location and contextual information through adaptive learning.
\end{abstract}

\section{Introduction}

In recent years, object detection has gained noteworthy improvements with the development of convolutional neural networks. Object detection is a computer technology related to computer vision and image processing which deals with detecting instances of semantic objects of a certain class (such as humans, buildings, or cars) in digital images and videos \citep{jiao2019survey}. Both in academic research and practical applications, like robotic vision, monitoring security, autonomous driving etc., object detection plays a vital role in computer vision tasks. In remote sensing, images or videos that need to be processed are captured by satellites or high-altitude flights, which makes OoIs (object-of-interest) too small to detect. Also the everyday scenarios have a variety of small objects. As in MS COCO dataset, small size objects account for a large portion of all instances, with instances less than 4\% of the image size accounting for nearly 30\% of all instances. Small objects are harder to detect than large ones, so we need to design a dedicated detector for small objects.

In general, object detectors can be divided into two categories: region proposal-based detector and regression-based detector. Region proposal-based detectors, such as R-CNN series \citep{rcnn,fast_rcnn,faster_rcnn,mask_rcnn}, SPP-net \citep{he2015spatial} and RFCN \citep{22rfcn_dai2016r}, are two-stage detectors with higher accuracy but slower inference speed. Regression-based detectors, such as YOLO series \citep{yolo,yolov2,redmon2018yolov3}, SSD series \citep{ssd,dssd_fu2017dssd}, and RetinaNet \citep{retinanet}, are one-stage detectors with faster inference speed but lower accuracy. Both one-stage or two-stage detectors have a common step, that is, feature extraction. And a high quality feature extractor can provide better features for later classification and regression networks. Depending on different requirements, one can embed a high-quality feature extractor into one-stage or two-stage detectors. 

\begin{figure}
\centering
\includegraphics[width=0.8\linewidth]{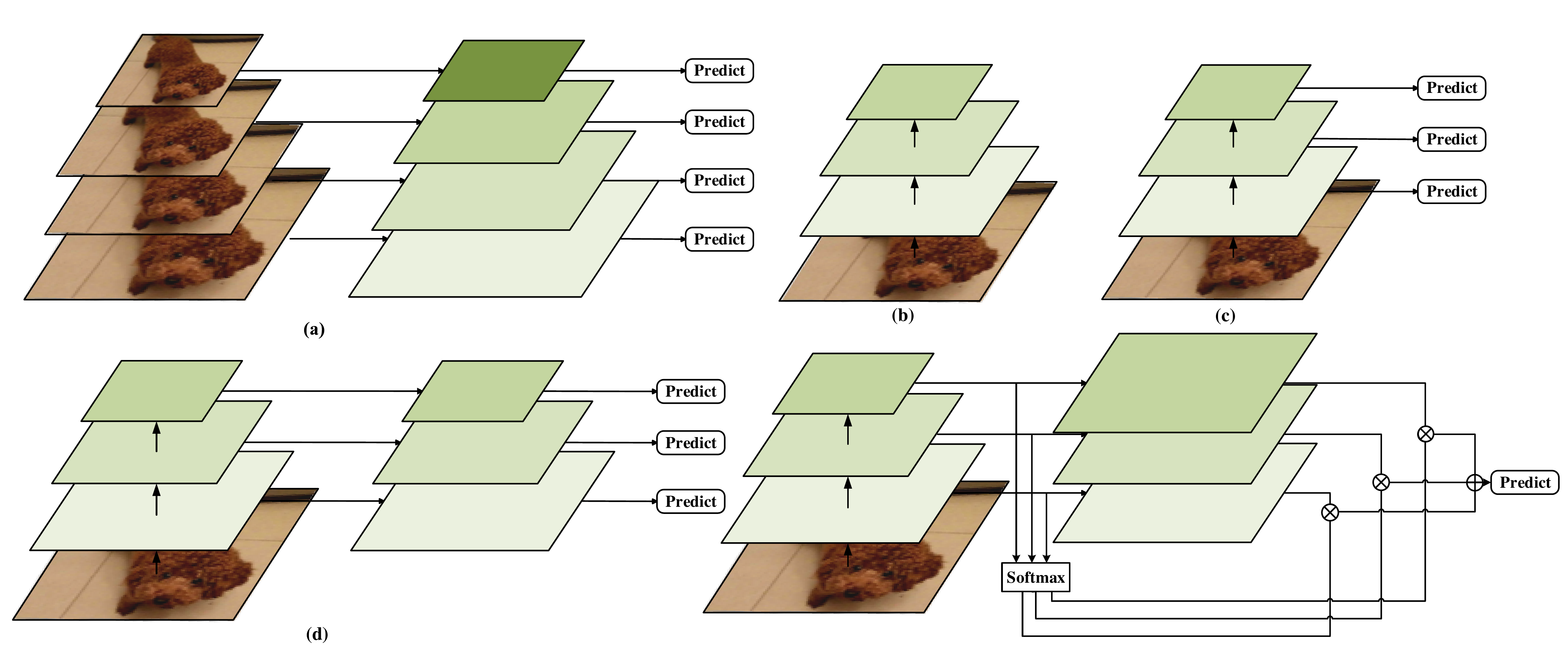}
\caption{Five methods utilize features for different sized object prediction. (a) Using an image pyramid to build a feature pyramid. Features are computed on each of the image scales independently, which is slow. (b) Detection systems \citep{fast_rcnn,faster_rcnn} use only single scale feature maps (the outputs of the last convolutional layer) for faster detection. (c) Predicting each of the pyramidal feature hierarchy from a ConvNet as if it is a image pyramid like SSD \citep{ssd}. (d) Feature Pyramid Network (FPN) \citep{fpn} is fast like (b) and (c), but more accurate. (e) Our MRAE architecture uses attention mechanism to select the most useful information to generate attention maps for small objects. In this figure, the feature graph is represented by a black-filled quadrilateral. The head network is represented by a black rectangle filled with "Predict". }
\label{fig_1}
\end{figure}

This paper designs a high-quality feature extractor for small object in various systems for detection and segmentation. A novel feature extraction method, called MultiResolution Attention Extractor (MRAE), adopts “attention” mechanism to focus on the most useful features. Our key motivation comes from the human vision system. When people look at a scene, they first focus on the most useful parts and ignore the rest, called the attention mechanism. Self-attention mechanism has been widely used in natural language processing (NLP) \citep{vaswani2017attention}. Recently, researchers have successfully migrated it to computer vision tasks \citep{wang2018non,zhang2018self,hu2018relation}. Self-attention, also known as internal attention, is an attention mechanism that connects different positions of a single image to calculate the image representation. Different from self-attention mechanism, some researchers use domain attention \citep{wang2019towards}, feature-based channel-wise attention mechanism \citep{hu2018squeeze} to handle different detection tasks. Although these methods have achieved good results in global object detection, they are not designed for small object detection. Our work focus on useful features and weighted-sum them, greatly enriching the features of small objects.

We choose Faster R-CNN as our base detector, which provide a basic two-stage detection procedure with high performance. We replace the feature extractor with our MRAE aiming to detect small objects. MRAE is simpler in design, plug and play, and can provide a high-quality feature extractor for small objects. Compared to the existing four feature-based approaches, MRAE is more effective at feature fusing (see Fig. \ref{fig_1}) because it automatically learns attention weight. 

Single-size feature map for small object detection is a plain method which does not distinguish objects of different sizes (Fig. \ref{fig_1}(b)). Therefore, it has the worst performance for small objects. In Faster R-CNN, the authors adopted a large stride 16 to get single size feature maps for classification. They suggest that even this large stride would yield good results, though accuracy might be further improved with a smaller stride. Image pyramid methods enlarge the original image to a larger size and extract features of each image scale independently (Fig. \ref{fig_1}(a)). Although input image is larger, the original small instance still has little information and features are not enriched. 

Using pyramidal feature hierarchy to distinguish different size objects, where given feature map level is responsible for specific scale of objects, like SSD (Fig. \ref{fig_1}(c)). On the one hand, however, the shallower layers of the backbone are not sufficient to recognize the category of objects. The classification sub-task requires semantic, abstract, and adequately processed features which are generated by high-level layers. On the other hand, the high-level layers produce low resolution features that lose a lot of location and edge information. Although the feature maps with the highest resolution lose the least location information and still retain the maximum feature maps in the case of small objects, after several subsequent convolutional layers, the semantic information will be enriched, which is also very useful. Therefore, naturally, we can combine the high-level and low-level features in an effective way. A more effective method, Feature Pyramid Network, leverages the architecture as a feature pyramid in which the object detection tasks are performed independently at each level (Fig. \ref{fig_1}(d)). The top-down branches and lateral connection combine the location information and semantic information of small targets, but this architecture is too heavy for small object detection. Moreover, FPN only fuses the highest level and the lowest level features for small object detection.

\begin{figure}
\centering
\includegraphics[width=0.6\linewidth]{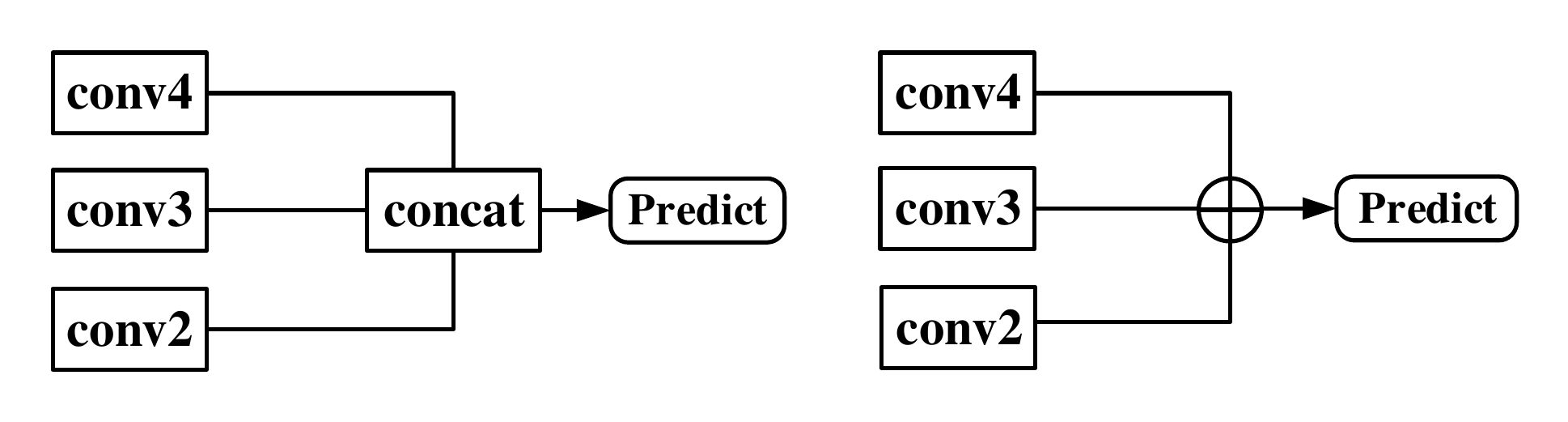}
\caption{Two architectures, Left: concatenating feature maps of several levels; Right: combining them through element-wise sum operation.}
\label{fig2}
\end{figure}

In contrast to these works, our MRAE highlights the most useful feature maps of several levels of the plain ResNet, also performs feature fusion to further enhance useful information (Fig. \ref{fig_1}(e)). The feature maps of several levels are weighted-sum, where the attention weights are learned by a small network (a convolutional layer, a fc layer, cosine similarity operation, and upsample operation), followed by a softmax layer. Now, the network forms new feature maps, called attention maps. This architecture is different from concatenating feature maps of several levels or combining them through element-wise sum operation (Fig. \ref{fig2}). The design of our MRAE is simple and preserves the advantages of feature pyramid network, which focuses on small object detection and is time efficient. We demonstrate the validity of MRAE on MS COCO object detection dataset \citep{5_10.1007/978-3-319-10602-1_48}. MRAE achieves a good AP (5.0\%), exceeding the baseline of 2.7\%, but with a faster convergence rate.  The improvement of experimental performance shows that this method has made some progress in small object detection and can be used in remote sensing vision task and other small object detection, tracking, and segmentation fields. Our main contributions can be summarized as follows:

(1) Soft attention is convenient, can backprop to learn where to attend. First, we propose a feature extraction method based on soft attention theory and find its limitations.

(2) We design an attention-based feature interaction  \textbf{MRAE} for small object detection as the second feature extraction method and demonstrate its effectiveness.

(3) Finally, we discuss three different attention mechanism-based methods through a series of experiments.

In the following sections, we further illustrate the effectiveness of our proposed MRAE. This paper first summarizes state-of-the-art object detectors, small object detection methods, and describes the visual attention mechanism in Section \ref{2relatedwork}. In Section \ref{3fan}, we describe two proposed methods in detail. Then we report a series of well-designed experiments to verify the effectiveness of our methods in Section \ref{4experiments}. Finally, we summarize the paper in Section \ref{5conclusion}.

\section{Related work}
\label{2relatedwork}
\subsection{Object detection}
Object detection is a computer vision task, including object instance classification and localization. Unlike image classification, localization can be regarded as a regression problem. At first, in the past few decades, the ad hoc hand-crafted features (e.g. SIFT and HOG) were widely adopted in image object detection. But these traditional methods rely heavily on expert experience and only simulate the first processing stage of human vision system (Complex cells in V1 are the first cortical area in the primate visual pathway). After the emergence of R-CNN, convolutional neural networks (CNNs) have been used by almost all object detectors. Moreover, R-CNN has four stages in the training process, which simulates the hierarchical and multi-stage processes of human vision system. Deep learning methods are good at dealing with the problem that the output representation is quite different from the initial representation. For example, in an object detection network, the input is image, that is, pixel, while the output is target, with a long distance, deep learning methods can solve it well. Although deep learning-based methods are “black box” models and lack mathematical proof, their performance in object detection always far exceeds that of traditional methods. 

\subsection{Small object detection methods}
Small objects occupy only a small part of a large area. "Small" here has two meanings, small size and low resolution. Sometimes, the resolution of small objects is too low to detect them. In general, small objects contain the least information, resulting in the least features can be extracted from them. Therefore, it is a great challenge to accurately detect small objects in images.

To address this, the traditional image pyramid and filter pyramid methods are to detect small objects in a pyramid hierarchy. In tiny face detection, \citet{hu2017finding} used an enlarged region to better fit the features of pre-trained network, which was proved an effective way. They proposed an architecture contains three re-scaled input branches, respectively handling specific scale face detection. However, the enlarged image incurs additional computational costs and a longer time consumption. Exploiting contextual information \citep{bell2016inside,chen2016r,kong2016hypernet,cao2018feature} is another effective approach. \citet{the_elephant_in_the_room_williams2011elephant} conducted a series of experiments showed that the features outside the RoI (region of interest) affect the final detection result. These operations need to generate additional contextual information, while we design a feature extractor which can capture useful localization and contextual information through adaptive learning.

Using deep learning methods always requires big data to make the network easy to train. \citet{DBLP:journals/corr/abs-1902-07296} proposed small object data augmentation methods to provide sufficient samples for deep learning network training. At the same time, adopting appropriate training methods is another area that needs to be addressed. For example, the pre-trained model uses high resolution images as input, which is not suitable for enlarged low resolution ones. Generative adversarial network is good at generating fake distribution similar to the input distribution. It utilizes a generator to deceive the discriminator to achieve a Nash equilibrium.  \citet{li2017perceptual} used GAN to generate super-resolved representations for small objects, making it as easy to train small objects as it is to train large ones. In general, GAN is difficult to converge. 

\subsection{Visual attention mechanism}
\citet{desimone1995neural} have found that visual attention problems can be defined by two basic phenomena. The first basic phenomenon is limited capacity for processing information. Specifically, only a small amount of information available on the retina can be processed and used to control behavior at any given time. Subjectively, giving attention to one target leaves less available for others. The second one is selectively, that is, the ability to filter out unwanted information. Subjectively, one is aware of attended stimuli and is largely unaware of unattended ones. \citet{itti2001computational}  proposed that “A unique 'saliency map' topographically encoding for stimulus conspicuity over the visual scene has proved to be an efficient and plausible bottom-up control strategy.” Although the attention maps in our MRAE do not compute the gradient to become topographic maps which represents conspicuousness of scene locations, they focus attention on the most salient location. Accordingly, attention maps in our MRAE can be regarded as the “saliency map”. Some feature maps have priority for small objects, while the others can be ignored. So the attention maps have useful information (with high attention weight) and are filtered out of the opposite information (with low attention weight). Existing detectors simply sum up the different resolution characteristics without prioritizing the most useful information or minimizing the impact of conflicting information. 

\citet{borji2012state} introduced the concept of attention, a universal concept that encompasses all factors that influence the selection mechanism, whether it is context-driven bottom-up (BU) or expectation-driven top-down (TD). FPN is an attention model composed of a context-driven bottom-up (BU) pathway and an expectation-driven top-down (TD) pathway. Inspired by the human visual attention mechanism, we designed two new feature extractor based on this mechanism.

\section{MultiResolution Attention Extractor (MRAE)}
\label{3fan}
In order to simulate the visual attention mechanism of human eyes in feature extraction, we design a new feature extractor, which can extract the most useful information and reduce the conflict information as much as possible. Our MRAE can be commonly used in both regression-based and region-based detectors. Considering that the region-based detector requires a feature extractor in two stages and has a high accuracy, we mainly adopt the classical region-based method Faster R-CNN to verify the effectiveness of our proposed MRAE. In any plain feature extractors, the final features usually take the output of the last layer or any intermediate layer in the forward propagation network (“plain” here means no feature pyramid, only a forward propagation network). \citet{5_10.1007/978-3-319-10602-1_48} proposed that generally smaller objects are harder to recognize and require more contextual reasoning to recognize. In order to integrate contextual information without additional pre-processing operation, we consider designing a novel multiresolution feature extractor. To achieve visual attention mechanism in feature level, we propose two implementation methods. The first one called soft attention, as compared to hard attention, uses max-pooling layer to generate an attention value. The second method is to design an attention-based feature interaction small network, in which a template feature level is defined and the final attention maps are generated according to the cosine similarity between the template and other feature levels.

\subsection{Soft attention}

\citet{xu2015show} first presented deterministic “Soft” Attention as compared to Stochastic “Hard” Attention.   The “soft” is derived from the softmax operation. And the stochastic attention only selects one feature level randomly. In section 4, we describe the effects of these attention approaches in detail. We focus on soft attention feature extraction method in this part. 

\begin{table}

\caption{ResNet architecture}
\label{resnet}
\centering
\begin{tabular}{lll|lll}
\toprule
layer name &  level  & 101-layer &layer name &  level & 101-layer\\
\midrule
conv1 & -  & ${7}\times{7}$, 64, stride 2\\
\cmidrule(r){1-3}
&&${3}\times{3}$ max pool,2  \\
\cmidrule(r){3-3}
conv2\_x & $C_1$ &
$\begin{bmatrix}
$ ${1}\times{1}$, 64$ \\
$ ${3}\times{3}$, 64 $\\
$ ${1}\times{1}$, 256$
\end{bmatrix} \times{3}$&conv4\_x& $C_3$ &$\begin{bmatrix}
$ ${1}\times{1}$, 256$ \\
$ ${3}\times{3}$, 256 $\\
$ ${1}\times{1}$, 1024$
\end{bmatrix} \times{23}$\\ 
\midrule
conv3\_x& $C_2$&$\begin{bmatrix}
$ ${1}\times{1}$, 128$ \\
$ ${3}\times{3}$, 128 $\\
$ ${1}\times{1}$, 512$
\end{bmatrix} \times{4}$&conv5\_x&$C_4$&$\begin{bmatrix}
$ ${1}\times{1}$, 512$ \\
$ ${3}\times{3}$, 512 $\\
$ ${1}\times{1}$, 2048$
\end{bmatrix} \times{3}$\\

\bottomrule
\end{tabular}
\end{table}

\begin{figure}
\centering
\includegraphics[width=\linewidth]{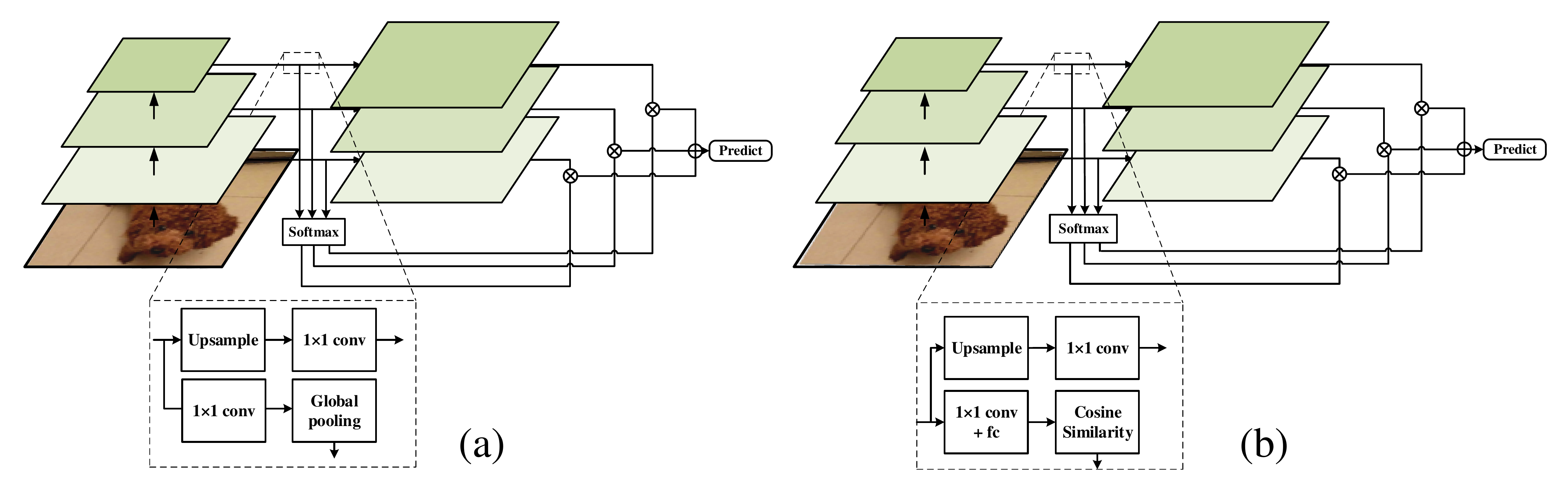}
\caption{The architecture of soft attention method and our MRAE. (a) Soft attention-based method. (b) Attention-based feature interaction MRAE. }
\label{fig_3}
\end{figure}

In ResNets, the feature extraction network has four levels, in which a level is composed of several conv layers with the same sized feature maps. We represent the four levels conv2, conv3, conv4, conv5 as $C_1, C_2, C_3, C_4$, respectively (Table \ref{resnet}). As \citet{vgg_simonyan2014very} proposed that the conv5 level is used to simulate the function of the hidden fully connected layers in VGG16 network. We use the first three levels ($C_1, C_2, C_3$) to generate attention maps. 

We attach a small network to obtain attention weights. The attention weight is defined as the weight of the features of each level to generate the final attention map. This small network consists of a  ${1}\times{1}$ conv layer and a max-pooling layer. The  ${1}\times{1}$ conv layer is to reduce dimension, which is equivalent to the sum in the depth direction. We set the output dimension of this convolutional layer to 1. The global max pooling layer is used to generate the maximum value of the feature map, that is, to capture the most noteworthy feature pixel. So far, we have got three representative feature pixels of three levels respectively. Then we send them to a softmax layer to obtain a set of normalized values:
\[a^i=\frac{exp(f(F^i))}{\sum_{k=1}^{3}exp(f(F^k))}.\]
In words, for each level, $f(\cdot)$ denotes the function of the small network, a mapping from feature maps to a salient value. The outputs of $C_2$ and $C_3$ are both smaller than those of $C_1$. So we add an ${2}\times$ up-sampling layer in $C_2$ and a ${4}\times$ up-sampling layers in $C_3$. So far, the feature maps corresponding to the three levels have been converted to the same size. Then, the attention weight  $a^i (i=1,2,3)$ is multiplied by feature maps of the corresponding level. Since the output depths of the three levels are not equal ($C_1:256, C_2: 512, C_3: 1024$), we add channels of the shallower levels ($C_1, C_2$) by a  ${1}\times{1}$ conv layer, so that the number of channels in the three levels is the same. Finally, the weighted feature maps of three levels are summed up element-by-element to obtain the final attention map:
\[A=\sum_{i=1}^{3}a^ig(F^i), g(\cdot): {1}\times{1} conv.\]

\subsection{Attention-based feature interaction MRAE}
Let the output of $C_t$ as template, we compute the cosine similarity of the template with the output of $C_{i(i!=t)},i=1,2,3$ respectively. For example, using $C_1$ as template, in $C_1$, $C_2$, and $C_3$, we append a small network composed of a ${1}\times{1}$ conv layer and a fc layer respectively to map to a vector for similarity calculation (Using $C_2$ or $C_3$ as template is straightforward and works well). We use cosine similarity
\[D^i[a][b]=\cos(F^t[a],F^i[b])=\frac{{F^t[a]}\cdot{F^i[b]}}{{|{F^t[a]}|}\cdot{|{F^i[b]}|}},\]
\noindent where $D^i[a][b]$ represents the similarity between vector $a$ and vector  $b$ ($i=1,2,3$). The cosine similarity between template feature map $F^t[a]$ and $F^1[b]$ indicates the attention degree of $C_1$. In the same way, the cosine similarity between $F^t[a]$ and $F^2[b]$ indicates the attention degree of $C_2$. To evaluate the attention weight $a^i$, we send $\{D^1=1,D^2, D^3\}$ to a softmax layer to obtain normalized value by
\[a^i=\frac{\exp(D^i)}{\sum_{k=1}^3 \exp(D^k)}.\]
Since small objects usually have a small number of pixels in the original image, their information will be gradually lost after several down-sampling layers. The feature map of $C_1$ has the largest size and retains the most information relative to the higher level. And the output sizes of the three levels are different, the feature maps of $C_2$ and $C_3$ are sent into a ${2}\times$ and ${4}\times$ up-sampling layers respectively. Finally, we add channels of the shallower levels ($C_1, C_2$) by a  ${1}\times{1}$ conv layer. We represent attention map  as the output of MRAE by
\[A=\sum_{i=1}^3 a^i F^i.\]

\section{Experiments}
\label{4experiments}

\begin{figure}
\centering
\includegraphics[width=0.6\linewidth]{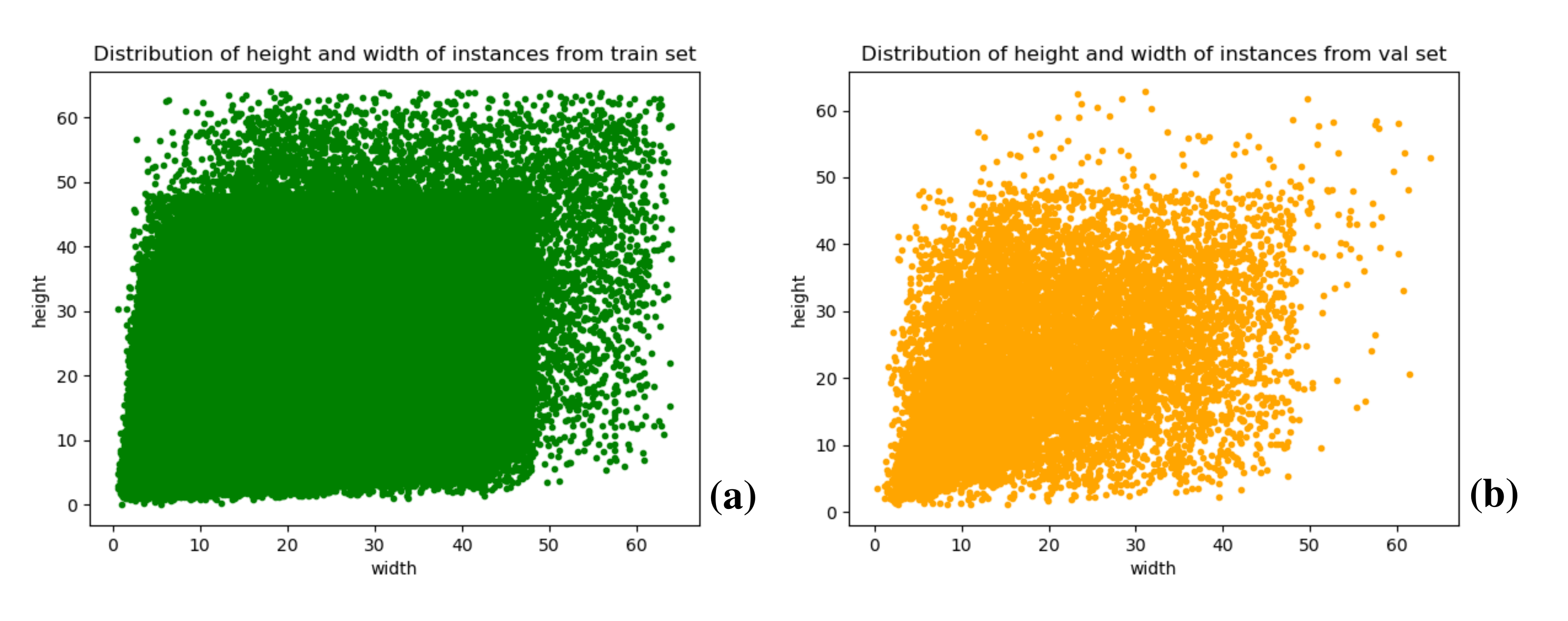}
\caption{The width and height distribution of all instances in train set (a) and validation set (b).}
\label{fig_4}
\end{figure}

We make a subset of MS COCO object detection dataset\footnote{http://cocodataset.org/\#download} which only contains small objects. Those selected small objects have bounding box area lower than ${32}\times{32}$. We divided the dataset into two parts, including 52032 images in the training set and 2164 images in the validation set.
We use clustering analysis to find some clusters of object scale and aspect ratio. Then we obtain four scales and three aspect ratio as our anchor parameter settings. The width and height distribution of all instances in train set and validation set is shown in Fig. \ref{fig_4}.

\textbf{Implementation details:}
We use batchsize = 1 because one NVIDIA GEFORCE 1080 (8 GB) only accommodate one image in the forward propogation and we use one GPU to train all the detectors. For input image size, the baseline training permits the large image size, ${1024}\times{600}$ and large NMS parameters (max detections per class: 300, max total detections: 600). For fair comparisons with the original Faster R-CNN and simulating hard attention method with randomly selected feature level, we run three baselines using the single-scale map of $C_1$ (inference speed: 22ms/image), $C_2$ (33ms/im) and $C_3$ (32ms/im). To train the soft attention method, MRAE, and hard attention baselines with the same hyperparameters, we adopt a smaller input size, ${512}\times{300}$, and small NMS parameters (max detections per class: 100, max total detections: 300). The initial learning rate is 0.0003 in the first 600k iteration, declined to 0.00003 in the following 300k iteration, and finally reached 0.000003. The momentum optimizer value is set to 0.9. All network backbones are pre-trained on the ImageNet1k classification set and then fine-tuned on the MS COCO small object dataset. We use the pre-trained ResNet-101 \citep{slim} model that is publicly availble\footnote{https://github.com/tensorflow/models/tree/master/research/slim\#Pretrained}. 

\textbf{Comparisons with baselines:} 
We report COCO-style AP and AR of the baseline \citep{huang2017speed}, soft attention method and our effective MRAE in Table \ref{table2}. For all experiments we use ResNet-101 as backbone, our soft attention extractor or MRAE constructed on top. For MRAE, we use the output feature maps of three levels as template respectively to find out which level has more important information. When we trained the soft attention method, the AP on the validation set barely grew. We conclude that the single general maxpooling operation result can not represent the corresponding feature level well. Using a larger input image size and a larger training step, the baseline gets 2.3\% AP. Our MRAE can grow AP rapidly and get 5.0\% AP with a 1/4 size input and less training steps. MRAE with template 2, which is $C_2$, the middle level we used got the highest AP with inference speed of 16ms per image (62.5fps). Thus, we find that the middle level $C_2$ has rich edge and semantic information, which is beneficial for small instance. All three feature levels as template achieve high performance, which validate the effectiveness of MRAE. Their accuracy has increased rapidly, which indicates that MRAE can extract more useful information for small objects automatically.

\textbf{Ablation studies:} 
We adopt mixed training to discover the effects of template feature level. After training Faster R-CNN+MRAE with template 1 reaching AP of 3.6\%, we continue to train it with template 2. When the training step reached 1310k, AP remained at 3.5\%, as in Table \ref{table3}.
This AP is much smaller than a pure MRAE using template 2 (5.0\%).  At the same time, it reduced the AP of pure Faster R-CNN+MRAE with template 1 by 0.1\%. Therefore, we found that template plays an important role in MRAE training. And the correlation between different feature layers is small. When they are combined effectively, they work better and more useful information for small objects is extracted.

\begin{table}
\caption{AP(top) and AR(bottom) of the baseline (Faster R-CNN ($C_1$), one of the hard attention methods), soft attention method, and our MRAE with three different parameter settings on the subset of  MS COCO object detection validation dataset. The image size is the input image size sent to the detector. The steps is the training steps to get the following AP and AR. AP: $AP^{0.5:0.95}$. The subscript of AR denotes the maxDets = 1,10,and 100. For example,  $AR_{1}$ denotes $(AR) @[ IoU=0.50:0.95\mid area = all\mid maxDets = 1 ]$. Because the dataset does not have large instances, the  $AP^{L}$ and $AR_{100}^{L}$ is ignored. The ms column denotes inference time per image.}
\label{table2}
\centering
\begin{tabular}{lllllllll}
\toprule
 & image size& steps & $AP$ &  $AP^{0.5}$ &  $AP^{0.75}$ &  $AP^{S}$ &  $AP^{M}$ &ms\\
\midrule
baseline $C_1$ &  ${1024}\times{600}$  & 1555.7k &2.3&4.5&1.9&2.3&3.6&-\\
baseline $C_1$ &  ${500}\times{300}$  & 1000k &2.3&5.0&1.6&2.3&4.7&22\\
baseline $C_2$ & ${500}\times{300}$  & 1046.6k&2.6&5.7&1.5&2.4&4.3&33\\
baseline $C_3$ & ${500}\times{300}$  &1009k&2.4 &6.5&1.1&2.3&5.8&16\\
Soft Attention & ${500}\times{300}$ & 1000k&0.04&0.1&0.006&0.06&0.05&-\\

MRAE template1 & ${500}\times{300}$ &1000k& 3.6& 7.4&2.9&3.6&6.5&16\\

MRAE template2 & ${500}\times{300}$ &1100k&\textbf{5.0}&\textbf{9.5}&\textbf{4.8}&\textbf{4.8}&\textbf{6.9}&16\\

MRAE template3 & ${500}\times{300}$ &1007.3k & 3.6& 7.3&3.2& 3.6 &6.4&14\\
\bottomrule
\toprule
 & image size& steps & $AR_{1}$ &  $AR_{10}$ & $AR_{100}$ & $AR_{100}^{S}$ &  $AR_{100}^{M}$\\
\midrule
baseline $C_1$ &  ${1024}\times{600}$  & 1555.7k &5.3&10.5&12.4&11.6&19.3\\
baseline $C_1$ &  ${500}\times{300}$  & 1000k &5.8&10.7&11.7&10.7&17.6\\
baseline $C_2$ & ${500}\times{300}$  & 1046.6k& 5.5&10.2&11.6&10.6&17.7\\
baseline $C_3$ & ${500}\times{300}$  &1009k&5.4& 10.0&11.3&9.9&19.7\\
Soft Attention & ${500}\times{300}$ & 1000k&0.01&0.3&1.1&1.1&1.3\\

MRAE template1 & ${500}\times{300}$ &1000k& 7.7& 14.6&\textbf{16.6}&15.7&22.6\\

MRAE template2 & ${500}\times{300}$ &1100k&\textbf{8.2}&\textbf{14.9}&\textbf{16.6}&\textbf{16.1}&20.8\\

MRAE template3 & ${500}\times{300}$ & 1007.3k& 8.1& 14.4& 16.3& 15.5 &\textbf{22.7} \\
\bottomrule
\end{tabular}
\end{table}

\begin{table}
\renewcommand{\arraystretch}{1.1}
\caption{AP (top) and AR (bottom) of Faster R-CNN+MRAE with template 1, Faster R-CNN+MRAE with template 2, and mixed training (Faster R-CNN+MRAE with template 1 and 2) on the subset of MS COCO object detection validation dataset.}
\label{table3}
\centering
\begin{tabular}{llllllll}
\toprule
 & image size& steps & $AP^{0.5:0.95}$ &  $AP^{0.5}$ &  $AP^{0.75}$ &  $AP^{S}$ &  $AP^{M}$\\
\midrule
MRAE template1 & ${500}\times{300}$ &1000k& 3.6& 7.4&2.9&3.6&6.5\\

MRAE template2 & ${500}\times{300}$ &1100k&\textbf{5.0}&\textbf{9.5}&\textbf{4.8}&\textbf{4.8}&\textbf{6.9}\\

MRAE temp 1+2 & ${500}\times{300}$ & 1310.7k& 3.5& 7.6& 2.6& 3.6 &5.7 \\
\bottomrule
\toprule
 & image size& steps & $AR_{1}$ &  $AR_{10}$ & $AR_{100}$ & $AR_{100}^{S}$ &  $AR_{100}^{M}$ \\
\midrule
MRAE template1 & ${500}\times{300}$ &1000k& 7.7& 14.6&\textbf{16.6}&15.7&22.6\\

MRAE template2 & ${500}\times{300}$ &1100k&\textbf{8.2}&\textbf{14.9}&\textbf{16.6}&\textbf{16.1}&20.8\\

MRAE temp 1+2 & ${500}\times{300}$ & 1310.7k& 7.3& 13.7& 15.6& 14.4 &\textbf{22.8}\\
\bottomrule
\end{tabular}
\end{table}

\begin{figure}
\centering
\includegraphics[width=\linewidth]{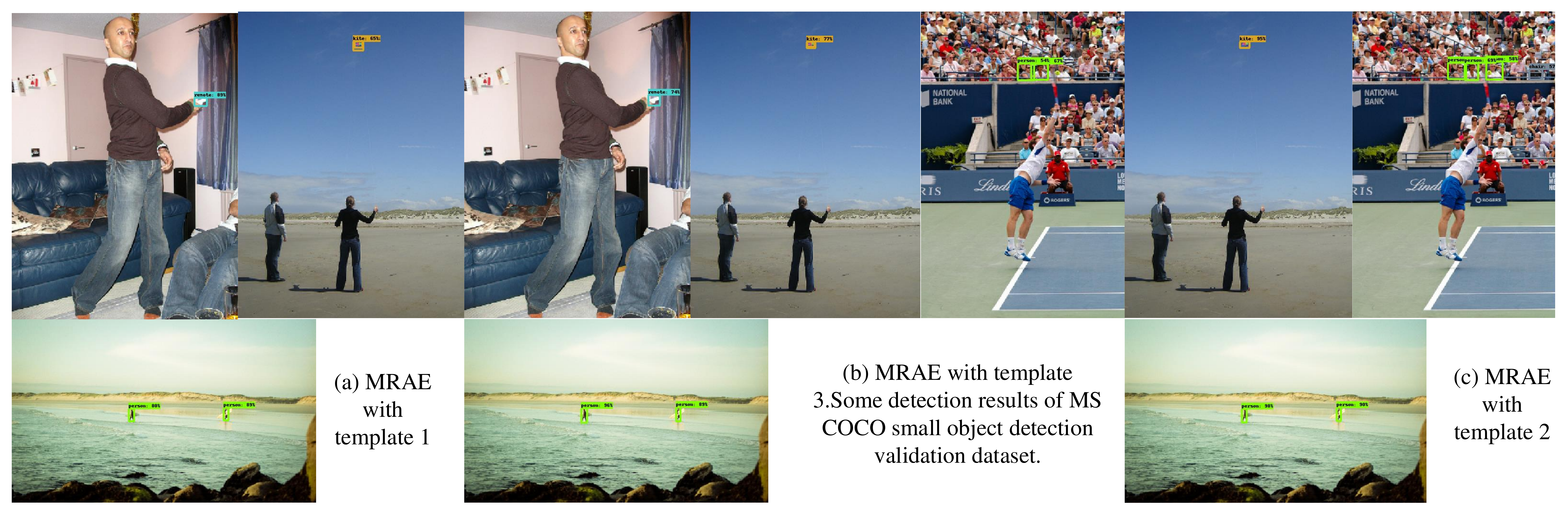}
\caption{Some detection results of MS COCO small object detection validation dataset.}
\label{fig_5}
\end{figure}
\section{Conclusion}
\label{5conclusion}
In this work, we propose a simple and effective MRAE for small object feature extraction based on human vision "attention" mechanism. First, we design a soft attention method, which we find difficult to converge in the experimental training process. Then, we propose a novel attention-based feature interaction small network. We demonstrate its efficacy by making a small object detection subset of MS COCO and report a series of experimental analysis. Our MRAE far exceeds the powerful baselines and is highly time efficient. Therefore, it provides a practical solution for research and application of multiresolution feature extraction without preprocessing the image or using GAN.

\section*{Broader Impact}

MRAE can be used in detection, segmentation, and tracking of small object scenarios. The theoretical basis is human visual attention mechanism. Our research can deepen the understanding of human visual attention mechanism. Using this extractor can bring some advantages, such as feature enhancement, features automatic extraction, low computational cost, no additional hyperparameter. 

We recommend that researchers understand the impact of using the MRAE in specific real-world scenarios. If the system fails, it will not get the desired automatic detection results. This method is only trained and proved to be effective in a small object environment, and the accuracy cannot achieve the commercial purpose. Please be careful to use it in areas requiring high safety coefficient.

The data used in the experiment were all images with an area less than $32\times32$ small targets in the public large data set. No manual screening was conducted, which ensured the fairness of the data.

\begin{ack}
The author thanks the researchers who have given guidance and the provision of the experimental server to her. And the author thanks her parents for their support and encouragement.

This work was supported in part by the State Key Program of National Natural Science of China (No.61836009).
\end{ack}

\small
\bibliographystyle{plainnat}

\bibliography{neurips_2020.bib}

\end{document}